\documentclass{article}

\usepackage{microtype}
\usepackage{graphicx}
\usepackage{natbib}
\usepackage{enumitem}
\usepackage{makecell}
\usepackage{subcaption}
\usepackage{array}
\usepackage{booktabs} 

\usepackage{hyperref}

\usepackage[preprint]{icml2026}

\usepackage{amsmath}
\usepackage{amssymb}
\usepackage{mathtools}
\usepackage{amsthm}

\usepackage[capitalize,noabbrev]{cleveref}

\theoremstyle{plain}

\theoremstyle{definition}

\theoremstyle{remark}

\usepackage[textsize=tiny]{todonotes}

\icmltitlerunning{Polaris: Scaling Up Instruction-Guided Image Generation Towards Millions of Personalized Style Needs}

\begin{document}

\twocolumn[
  \icmltitle{Polaris: Scaling Up Instruction-Guided Image Generation Towards Millions of Personalized Style Needs}



\icmlsetsymbol{equal}{*}

\begin{icmlauthorlist}
  \icmlauthor{Zhi-Kai Chen}{sai,nkl}
  \icmlauthor{Jun-Peng Jiang}{sai,nkl}
  \icmlauthor{Jun-Jie Tao}{nju}
  \icmlauthor{De-Chuan Zhan}{sai,nkl}
  \icmlauthor{Han-Jia Ye}{sai,nkl}
\end{icmlauthorlist}
\icmlaffiliation{sai}{
  School of Artificial Intelligence, Nanjing University, China
}
\icmlaffiliation{nkl}{
  National Key Laboratory for Novel Software Technology, Nanjing University, China
}
\icmlaffiliation{nju}{
  Nanjing University, China
}
\icmlcorrespondingauthor{Han-Jia Ye}{yehj@lamda.nju.edu.cn}

  \icmlkeywords{Machine Learning, ICML}

{%
\renewcommand\twocolumn[1][]{#1}%
\begin{center}
    \centering
    \captionsetup{type=figure}
    \vspace{-0.1cm}
    \includegraphics[width=1\textwidth]{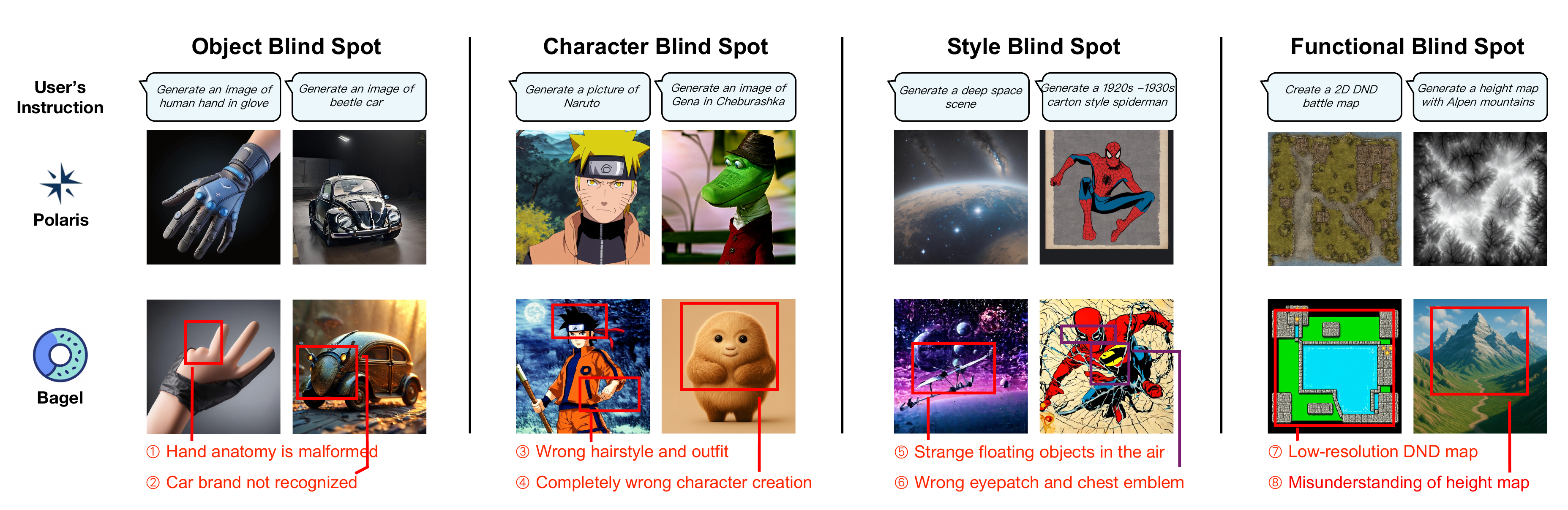}
    \vspace{-0.6cm}
    \captionof{figure}{Although large models fine-tuned on trillions of tokens better understand
    diverse user instructions, such as style generation, they still exhibit significant blind spots.
    We introduce Polaris, a retrieval framework over the Stable Diffusion model library that automatically
    identifies and invokes the most relevant models, addressing these blind spots and enabling diverse,
    personalized image generation.}
    \label{blind_spot}
\end{center}%
}

  \vskip 0.3in
]



\printAffiliationsAndNotice{}  

\begin{abstract}
Users increasingly expect image generation models to quickly adapt to highly diverse and personalized requirements, such as producing images with distinctive styles or characteristics. Traditional approaches rely on fine-tuning, which is costly and difficult to scale. To cope with these limitations, the community has accumulated a growing library of fine-tuned modules and adapters, where each component targets specific generation needs and collectively serves as a foundation for handling new demands.
This naturally raises a question:  instead of repeatedly training new models, can we systematically exploit this expanding ecosystem to better fulfill user instructions? To this end, we present Polaris, an intelligent retrieval framework that automatically selects and integrates suitable models from the model library based on a user's instructions.
The key insight is that harnessing such a massive and heterogeneous pool requires not only finding the most relevant modules among thousands of candidates, but also aligning them effectively for instruction-driven generation and editing. 
Polaris addresses this challenge by indexing over 6,500 checkpoints and 75,000 adapters, and retrieving the most relevant components given a user's input and instruction. In doing so, it delivers scalable, controllable, and well-aligned generation---without any additional training.
\end{abstract}

\section{Introduction}
\label{sec:intro}
The field of image generation has witnessed remarkable progress~\citep{diff_base_img_gen_survey,personalized_img_gen_sursey,t2i_t2v_survey}, particularly with the emergence of diffusion models that demonstrate strong capabilities in producing high-quality, diverse images~\citep{diffusion,ldm,sdxl,diffusion_survey}. These models have been successfully applied to a wide range of practical downstream tasks, including super-resolution~\citep{super}, image completion~\citep{complete}, style transfer~\citep{transfer}, image segmentation~\citep{segment}, and image editing~\citep{edit}. Such advances have made generative models increasingly accessible to end users.

Users’ needs are often highly diverse~\citep{Varif_ai} and deeply personalized with real-time feedback. For example, they may wish to generate images in distinctive artistic styles, replicate niche visual domains, or perform complex editing operations on existing content. Yet, due to issues such as data imbalance~\citep{class_imbalance_survey} and catastrophic forgetting~\citep{catastrophic_survey,overcome_catastrophic,continual_learning}, no single model can reliably master this broad spectrum of styles and tasks. Even large-scale and heavily fine-tuned foundation models still fail to handle many of these long-tail requirements, despite being trained on billions of examples. For precise user needs, small-scale finetuning or generating task-specific adapters such as LoRA~\citep{lora} often provides far more accurate customization than relying solely on a single foundation model. Yet each new requirement still necessitates training a new adapter or checkpoint, which incurs non-negligible costs in data collection and tuning.

Building on this context, it is important to recognize that many users have already adopted more lightweight strategies in practice. Instead of training models from scratch, which is costly and time-consuming, they often start from a base model and leverage community-released checkpoints or adapters that capture distinctive styles, domains, or editing capabilities. Such components are openly shared through platforms including Civitai, PixAI, and Tensor.art, which collectively host tens of thousands of publicly available resources. This ecosystem substantially reduces the cost of customization and has enabled the rapid prototyping of diverse generative workflows. Motivated by these observations, we ask whether these community-contributed modules can be systematically harnessed to extend and promote the capabilities of our model. If feasible, this paradigm would provide a lightweight and scalable alternative to traditional personalization methods, while simultaneously raising important challenges in component selection and integration for downstream tasks such as instruction-driven generation.

However, systematically exploiting this rich and heterogeneous ecosystem is far from trivial. Two core challenges arise. First, user queries for personalized image generation are often semantically complex and multimodal. A system must not only interpret open-ended instructions and visual references, but also accurately ground them to the underlying style or domain that best reflects the user’s intent. Second, as the pool of checkpoints and adapters grows to tens of thousands, retrieval must be efficient and scalable; naïve search quickly becomes impractical for interactive generation. To address these challenges, we introduce Polaris, a unified framework for instruction-driven style and capability selection. Polaris formulates this process as a multimodal retrieval task: it maps user intent to appropriate community models across style, domain, and concept dimensions. It incorporates an efficient search-and-rerank pipeline that ensures both
scalability and responsiveness at interactive latency.

With these designs, Polaris systematically harnesses community-contributed checkpoints and adapters to deliver instruction-driven image generation without requiring any additional training. It can directly parse user instructions, recommend the most suitable models, and produce high-quality outputs. Compared with approaches that optimize solely for instructional input, Polaris achieves substantial gains by leveraging its large-scale model library. As shown in Fig.~\ref{blind_spot}, when large pretrained models struggle with rare or specialized concepts, Polaris offers an alternative generation route that covers these blind spots through targeted community adapters, while maintaining clear advantages in inference efficiency. Together, these results establish Polaris as a scalable and effective paradigm for controllable image generation and editing. Our contributions can be summarized as follows:

\begin{itemize}[noitemsep,topsep=0pt,leftmargin=*]
\item We emphasize the need to address highly diverse, user-specific requirements while maintaining efficiency, and demonstrate that leveraging a large model library offers a practical and flexible solution.

\item Our Polaris retrieves models using combined text and image queries and incorporates an efficient reranking strategy, enabling fast and scalable selection from tens of thousands of candidates.

\item By harnessing the model library, Polaris delivers high-quality outputs that adapt to diverse user instructions with real-time performance, enabling scalable and customizable image generation.
\end{itemize}

\paragraph{Remark.} 
Polaris focuses on a style-aware instruction setting, where the central requirement is to interpret and realize stylistic intent. This differs from the conventional image-editing setting, which centers on localized content modifications, whereas Polaris enables style-driven transformations and style-consistent generation, including capabilities beyond standard editing.

\section{Related Works}
\label{sec:related}
\subsection{Diffusion-Based Text-to-Image Generation.}
Diffusion models generate images by progressively denoising random noise under the guidance of a learned score function~\citep{ddpm,score_based_generative}, which enables high-quality and diverse synthesis. Stable Diffusion~\citep{ldm} further introduced the latent diffusion framework, operating in a compressed latent space to achieve an efficient tradeoff between computational cost and image fidelity. Since its initial release, multiple versions (v1, v2, SDXL~\citep{sdxl}, and the recent SD3/3.5~\citep{sd3}) have been developed, continually improving resolution, fidelity, and controllability, making diffusion a dominant paradigm for text-to-image generation.

\subsection{Model Adaptation for Personalized Generation}
Model adaptation for personalization can be divided into checkpoint-based and adapter-based methods~\citep{view_para_efficient,para_efficient_survey}. The former, such as full fine-tuning~\citep{finetune} and DreamBooth~\citep{dreambooth}, directly update model parameters to learn new concepts, producing new checkpoints but often at the cost of data efficiency and forgetting. The latter, including LoRA~\citep{lora} and Hypernetworks~\citep{hyper}, insert lightweight adapters that preserve the base weights and support modular, efficient customization. In general, checkpoint-level adaptation enables stronger style or domain shifts, while adapter-based methods are better suited for fine-grained or localized edits.

Early model library~\citep{zoo_2021,zoo_vit} were constructed at the checkpoint level, collecting full model states for reuse and comparison. With the growing adoption of adapters such as LoRA, recent work has shifted toward adapter-level library~\citep{lorahub,loraretriever,stylus}. However, these approaches leave several gaps: they do not establish a clear index linking checkpoints and their associated adapters, they only support retrieval at a single granularity (either checkpoints or adapters), and in the image generation domain (e.g., Stylus) adapter retrieval is restricted to a single modality without support for multimodal search.

\begin{figure*}[ht]
\centering
\vspace{-3pt}
\includegraphics[width=\linewidth]{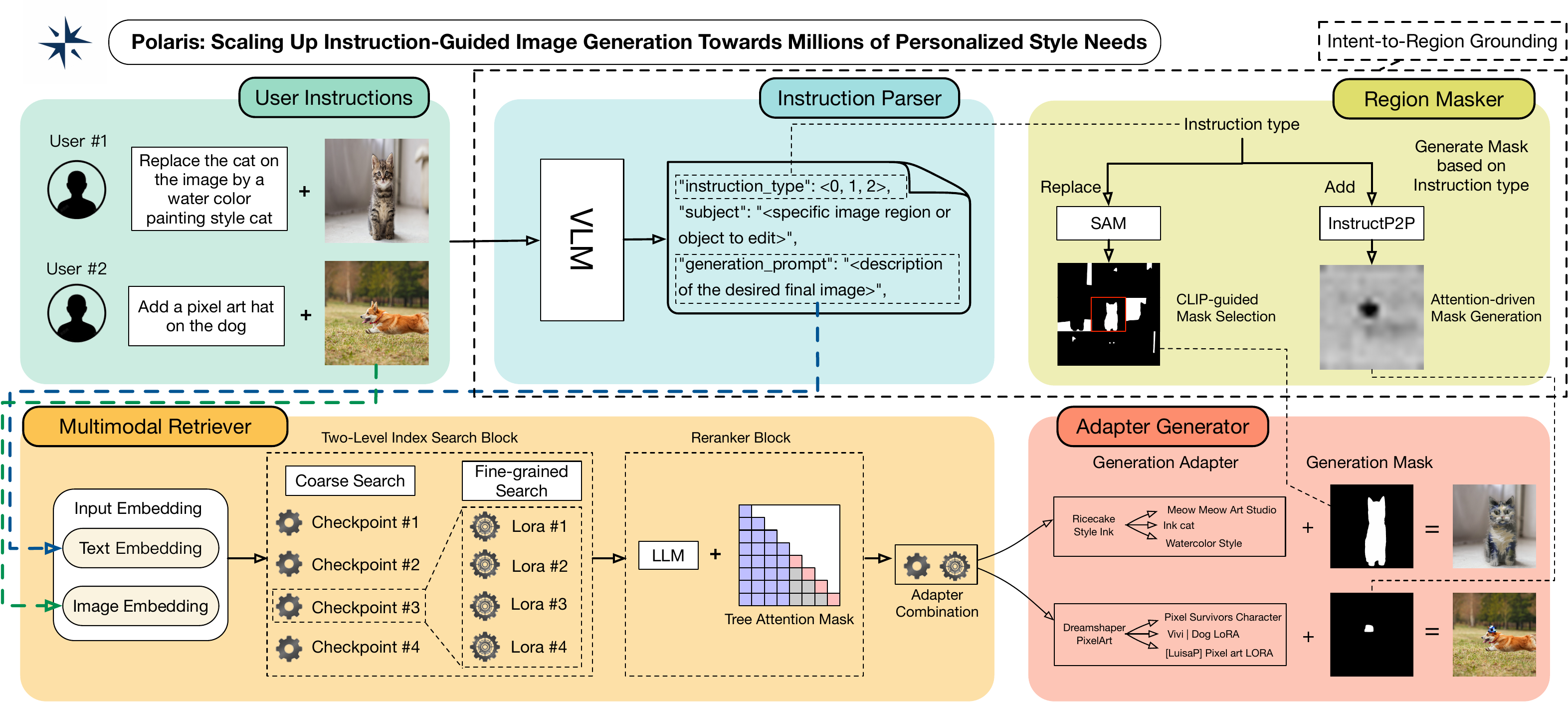}
\caption{
An overview of Polaris for instruction-guided model selection. The method is divided into two parts: (1) an Instruction Parser and Region Masker for adapting to image editing, and (2) a Multimodal Retriever for finding suitable checkpoints and LoRA adapters based on user requirements. The retrieved adapters and generated masks are combined to guide diffusion-based image generation.}
\label{method_overview}
\vspace{-10pt}
\end{figure*}

\section{Preliminaries}

In this section, we first revisit the conventional instruction-guided image generation task and discuss representative approaches. We then introduce our extended formulation with personalized needs, which targets diverse and user-specific requirements and introduces new challenges.

\subsection{Instruction-Guided Image Generation}
Instruction-guided image generation aims to produce images that follow both a given text instruction $I$ and, optionally, a reference image $T$. A common paradigm for this task is based on Stable Diffusion (SD), where the generation process is controlled by a model configuration. Typically, such a configuration includes a checkpoint $c$, which defines the base generative capability, and a LoRA adapter $l$, which provides task- or style-specific adaptations corresponding to users' needs.

To better align generative models with user intent, existing methods mainly take two directions: enhancing diffusion models with instruction-following ability, and leveraging multimodal large language models (MLLMs)~\citep{mllm_survey,mmllms_survey} to unify reasoning with generation. For instance, InstructP2P~\citep{instructp2p} fine-tunes Stable Diffusion on synthetic instruction–image pairs, enabling edits guided directly by natural language.

More recent work extends LLMs or MLLMs to incorporate image synthesis within a unified framework~\citep{unified_survey}. These approaches either model text and images jointly in an autoregressive manner (e.g., Emu~\citep{emu}, LaVIT~\citep{lavit}, Chameleon~\citep{chameleon}), or combine autoregressive reasoning with diffusion-based generation (e.g., Transfusion~\citep{transfusion}, LMFusion~\citep{lmfusion}). Such unified designs represent a promising step toward more general-purpose foundation models for controllable image generation with user intent.

\subsection{Instruction-Guided Image Generation with Personalized Needs}
While existing methods achieve good performance, directly fine-tuning models for each individual user is impractical in real-world scenarios. User requirements are inherently diverse, and a single monolithic model cannot adequately capture such variation with high efficiency. A more practical solution is to leverage the growing ecosystem of community-shared resources, including both base checkpoints and lightweight adapters for specific needs. We therefore extend the \emph{Model Library} beyond conventional checkpoints to also include adapters, enabling compositional model configurations. 

Formally, we define the model library $\mathcal{Z} = \mathcal{C} \times \mathcal{L}$, where $\mathcal{C} = \{ c_1, \dots, c_{N_c} \}$ denotes the set of base checkpoints and $\mathcal{L} = \{ l_1, \dots, l_{N_l} \}$ the set of LoRA adapters specialized for different styles, domains, or objects. In practice, we collect over 6,500 checkpoints and 75,000 adapters from Civitai~\footnote{\url{https://civitai.com}}, resulting in millions of possible model configurations for scalable and highly customizable image generation.

In this setting, our goal is to leverage the Model Library to adaptively match user requirements with suitable model configurations. Specifically, given a user query $q = (I, T)$, we aim to retrieve the most relevant checkpoint $c$ and adapter $l$ from $\mathcal{Z}$ and use them for generation or editing. The process involves three key steps: (1) obtaining a representation of the query that captures both textual and visual intent, (2) retrieving the checkpoint and LoRA adapter that best align with user’s style intent, and (3) performing image generation or editing with the selected model configuration.

\paragraph{Multi-Modal Instruction Embedding.}
To represent the query, we define an embedding $z_q$ that serves as the input to model retrieval. Rather than fusing $I$ and $T$ in a fixed manner, we select the dominant modality based on the type of instruction. Formally, we define:
\[
z_q = 
\begin{cases}
f_{\text{text}}(T), & \text{if } \beta(q) = \text{text-dominant}, \\
f_{\text{img}}(I),  & \text{if } \beta(q) = \text{image-dominant},
\end{cases}
\]
where $f_{\text{text}}$ and $f_{\text{img}}$ are the text and image encoders, and $\beta(q)$ denotes a modality selection rule defined over the instruction semantics.

This selection reflects a key assumption in our setting: different types of instructions emphasize different input modalities. For example, style transfer or global transformations rely more on $T$, while object-centric edits guided by image content rely more on $I$.

\paragraph{Retrieval Objective.}
Each candidate checkpoint $c \in \mathcal{C}$ and adapter $l \in \mathcal{L}$ has a precomputed embedding $\phi_c$, $\phi_l$. The similarity between the query and each candidate is computed as:
\[
s_c(q, c) = \langle z_q, \phi_c \rangle, \quad
s_l(q, l) = \langle z_q, \phi_l \rangle
\]
We select the most relevant model components by:
\[
c^* = \arg\max_{c \in \mathcal{C}} s_c(q, c), \quad
l^* = \arg\max_{l \in \mathcal{L}} s_l(q, l)
\]

The final image $x^*$ is synthesized by applying the selected model configuration $\Theta_{c^*, l^*}$ under the guidance of the instruction $T$:
\[
x^* = \arg\max_{x} p_{\Theta_{c^*, l^*}}(x \mid f_{\text{text}}(T))
\]
Through the model library paradigm, diverse user needs can be flexibly satisfied by selecting appropriate checkpoints and adapters. Nevertheless, two core challenges remain: (1) how to construct multi-modal embeddings that faithfully capture user intent under heterogeneous instructions, and (2) how to perform retrieval efficiently over a large repository of model components.

\section{Method}
\subsection{Overview of Polaris}
We aim to enhance both the stylistic accuracy and diversity of instruction-guided image generation within our model library, facing two key challenges. First, textual instructions can be structurally complex. To interpret them, we use an \emph{Instruction Parser} that processes text-only inputs to extract actionable intent, and a \emph{Region Masker} that provides spatial grounding when the instruction specifies localized changes. These components focus purely on instruction understanding. Second, we use a \emph{Multimodal Retriever} that performs checkpoint and adapter retrieval by jointly leveraging text and image embeddings, with an \emph{LLM Tree Rerank} module for reranking.
An overview is shown in Figure~\ref{method_overview}.

\subsection{Instruction Parser}
Since Polaris targets instruction-guided image generation but our retriever is primarily optimized for style personalization, it does not natively decompose free-form user instructions. Therefore, we introduce an \emph{Instruction Parser} to extract the essential intent from a user query $q = (I, T)$, where $I$ is an optional reference image and $T$ is the textual instruction, and convert it into a structured representation suitable for downstream retrieval and generation.

The parser leverages a vision–language model (VLM) to decomposes the instruction into three main elements, which we formally define as a mapping:
\begin{equation}
\mathcal{P}: (I, T) \mapsto (t_1, t_2, t_3)
\end{equation}
where $t_1$ denotes the \emph{instruction subject}, i.e., the target object or region to be modified or generated; $t_2$ denotes the \emph{instruction type}, specifying the operation (e.g., ``modify'', ``replace'', ``stylize''); and $t_3$ denotes the \emph{generation prompt} $T'$, a refined or reformulated version of $T$ designed to better align with the conditional diffusion model. In our implementation, we leverage Qwen2.5-VL-7B~\citep{qwen25vl} as the vision-language model to perform this instruction parsing. For a more detailed description of the prompts used by the instruction parser, please refer to the Appendix~\ref{app:IP}.

\subsection{Region Masker}
The Region Masker grounds the instruction subject $t_1$, identified by the Instruction Parser, onto the user-provided image $I$. Its role is to localize the target region that should be modified while preserving unrelated areas. For instructions that apply to the entire image (e.g., global style changes), the Region Masker is skipped, and the full image is treated as the modification region.

\paragraph{General Case ($t_2 =$ modify-type).} We first apply a pre-trained segmentation model to generate a set of candidate masks $\mathcal{M} = \{ m_1, m_2, \dots, m_{N_m} \}$. For each mask, we compute a relevance score using a value function $v(\cdot)$ that measures the semantic similarity between the mask region and the parsed subject. The optimal mask is selected as:
\begin{equation}
m^* = \arg\max_{m \in \mathcal{M}} v(m, t_1)
\end{equation}
The selected mask $m^*$ provides spatial constraints for the Adaptive Editor, ensuring that edits remain localized. In practice, we use the Segment Anything Model (SAM)~\citep{sam} to propose candidate regions, and compute relevance using a CLIP-based similarity function adapted to masked image inputs~\citep{ovseg}. Here, CLIP~\citep{clip} refers to a vision–language model trained to align images and text in a joint embedding space, making it well suited for evaluating mask–subject correspondence.

\paragraph{Special Case ($t_2 =$ add-type).}
For instructions involving object addition, SAM alone is insufficient. To better capture the intended region, we intersect the attention map from InstructP2P with the SAM-generated masks~\citep{Zone}, yielding a refined mask for the most relevant area to insert.

\begin{figure}[!ht]
    \centering
    \includegraphics[width=0.75\linewidth]{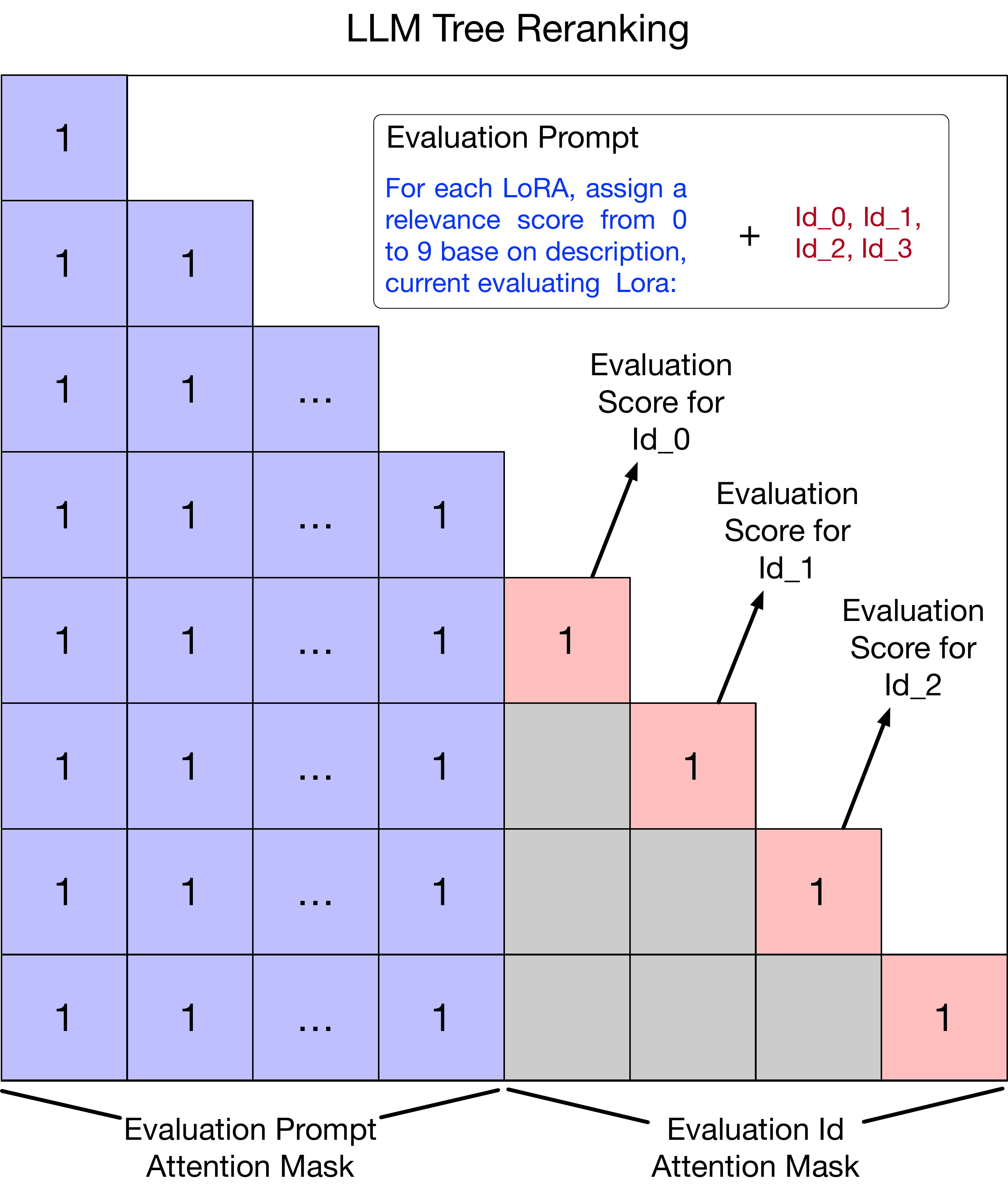}
    \caption{An overview of the tree rerank method. By modifying the attention mask of the LLM, we enable the model to evaluate multiple adapters simultaneously while generating only a single token per evaluation.}
    \label{fig:tree_rerank}
    \vspace{-10pt}
\end{figure}

\subsection{Multimodal Retriever}
We construct two modality-specific indices offline: one for checkpoints $\mathcal{C}$ and one for adapters $\mathcal{L}$. Each item—whether a base checkpoint or a LoRA adapter—is embedded into a shared representation space that combines textual and visual information using a CLIP encoder. 

For each model $x$ (checkpoint $c$ or adapter $l$), we first construct a visual embedding to capture its style-related generative behavior.
We leverage exemplar images produced by the community using the same model, as they implicitly encode its visual characteristics—such as artistic style, color palette, composition tendencies, and domain specificity.
Since user-uploaded content may include noise or examples not representative of the model’s core capability, we apply filtering and quality control to retain only high–semantic consistency samples.
The curated exemplars are then used to compute an average image embedding, grounding retrieval on the model’s actual observable output distribution.

Complementarily, we also extract a text embedding from associated metadata, including model name, tags, release notes, and user descriptions, which provides high-level semantic cues aligned with user intents. Formally:

\begin{equation}
\begin{aligned}
\mathbf{z}^{\mathrm{img}}_x  &= \frac{1}{m} \sum_{i=1}^{m} \mathrm{CLIP}_{\mathrm{img}}(I_{x,i})\\
\mathbf{z}^{\mathrm{text}}_x &= \mathrm{CLIP}_{\mathrm{text}}(\mathrm{metadata}_x)
\end{aligned}
\end{equation}
Both components are concatenated to yield the final multimodal representation:
\begin{equation}
\mathbf{z}_x = [\mathbf{z}^{\mathrm{text}}_x \,\Vert\, \mathbf{z}^{\mathrm{img}}_x] \in \mathbb{R}^{2d}
\end{equation}
The resulting embeddings are stored in the checkpoint index $\mathcal{C}$ and adapter index $\mathcal{L}$, enabling unified retrieval across modalities. To select with a user query $q$, we adopt a two-stage retrieval strategy:

\textbf{Level-1: Coarse checkpoint selection.}
We first identify the most relevant base model by computing similarity scores $s_c(q, c)$ between the query and all entries in  $\mathcal{C}$:
\begin{equation}
c^* = \arg\max_{c \in \mathcal{C}} \; s_c(q, c)
\end{equation}

\textbf{Level-2: Fine-grained adapter selection.}
Conditioned on the selected checkpoint $c^*$, we restrict the adapter search to a local neighborhood $\mathcal{L}_{c^*} \subseteq \mathcal{L}$ and identify the best-matched LoRA:
\begin{equation}
l^* = \arg\max_{l \in \mathcal{L}_{c^*}} \; s_l(q, l)
\end{equation}
The final pair $(c^*, l^*)$ is then used to generate the output image conditioned on the user instruction $T$, and optionally a reference image $I$. To further improve adapter selection, we incorporate an LLM-based refinement module inspired by Stylus~\citep{stylus}. The LLM decomposes user instructions into sub-tasks and allocates LoRA adapters, which provide more precise control over global style and object-level edits.

\subsection{LLM-Based Reranking}
In the \emph{Multimodal Retriever} stage, we use LLM to refine retrieved candidates based on textual compatibility with user instructions and model metadata, similar to Stylus’s reranking strategy. However, each LoRA adapter is associated with a textual description that must be provided to the LLM. As a result, when the candidate set is large, the prompt length can exceed 20000 tokens. This results in significant computational overhead due to the quadratic growth of LLM self-attention with sequence length. 

To address this, we propose a novel inference acceleration strategy called \textit{Tree Reranking}, which balances retrieval quality and computational cost, making the reranking process more efficient and practical for deployment. The key idea is to modify the LLM’s attention mask so that multiple branches (representing different adapter candidates) can be scored in a single forward pass.
As illustrated in Figure~\ref{fig:tree_rerank}, the reranker generates scores for multiple LoRAs simultaneously, then prunes low-scoring candidates and performs additional rounds of reranking on the remaining branches.
By progressively narrowing the candidate pool, we drastically reduce the effective input length required at each stage. This tree-based speculative reranking effectively reduces redundant computation in attention mechanism, which would otherwise scale non-linearly with sequence length. In practice, our Tree Reranking approach yields a substantial speedup for the reranking module while maintaining high-quality adapter selection.

\subsection{Adaptive Editor}
To enable flexible and precise image editing, our system integrates three distinct types of adapters:  
(1) a base checkpoint that governs the overall image style,  
(2) a style LoRA that modulates finer stylistic elements, and  
(3) an object-specific LoRA that controls the visual characteristics of specific entities (e.g., animal breeds or object categories).

The selection strategy for these adapters is conditioned on the nature of the user instruction. For instructions that focus on object replacement without explicit style control (e.g., “Replace the cat in the image with another breed”), we retrieve the overall style checkpoint and the fine-grained style LoRA from the input image to preserve visual consistency with the original scene, while extracting the object-specific LoRA based on the textual instruction. In contrast, for instructions that involve both object manipulation and explicit style transformation (e.g., “Replace the cat with a watercolor-style one”), we retrieve the overall style checkpoint and style LoRA from the textual prompt to reflect the desired stylistic shift, while still relying on the object-specific LoRA derived from the prompt to control entity-level appearance. Further details on how we distinguish between these instruction types and perform retrieval accordingly are provided in the appendix~\ref{app:AE}.

\begin{table*}[ht!]
\centering
\caption{User-Bench results on two subcategories: Local style change and Style extraction. VQ (Visual Quality), EQ (Edit Quality), and their geometric mean (Overall) are reported. Polaris achieves significant gains on local style change compared to InstructP2P and attains competitive performance with unified multi-modal models, while obtaining the best results on style extraction.}
\label{tab:ourbench_sub}
\begin{tabular}{lccccccccc}
\hline
 & \multicolumn{3}{c}{\textbf{Style extraction}}  
 & \multicolumn{3}{c}{\textbf{Local style change}} \\
\cmidrule(lr){2-4} \cmidrule(lr){5-7}
 & VQ  & EQ & Overall 
 & VQ & EQ & Overall \\
\hline
InstructP2P & 4.41  & 4.08  & 4.24 & 5.11 & 5.24 & 5.17   \\
BAGEL       & 6.86  & 4.27 & 5.41 & 6.20 & 6.81 & 6.50  \\
Polaris     & 7.33  & 5.50 & 6.35 & 5.84 & 6.19 & 6.01  \\
\hline
\end{tabular}
\vspace{-5pt}
\end{table*}

\begin{table*}[ht!]
\centering
\caption{The experimental results on Gedit-Bench~\citep{gedit} demonstrate the effectiveness of our strategy. Compared with InstructP2P, which relies on finetuning to support instructional inputs, our approach leverages retrieval from a model library to achieve superior performance.} 
\label{tab:eval}
\begin{tabular}{lccccc}
\hline
 & \makecell{\textbf{Background} \\ \textbf{change}} 
 & \makecell{\textbf{Material} \\ \textbf{alter}} 
 & \makecell{\textbf{Style} \\ \textbf{change}} 
 & \makecell{\textbf{Subject} \\ \textbf{add}} 
 & \makecell{\textbf{Subject} \\ \textbf{replace}} \\
\hline
InstructP2P & 3.70 & 3.39 & 4.60 & 3.18 & 3.80 \\
BAGEL& 7.06 & 6.40 & 6.13 & 8.06 & 6.71 \\
Polaris & 4.31 & 4.57 & 5.00 & 3.23 & 3.76 \\
\hline
\end{tabular}
\end{table*}
\section{Experiment}
\paragraph{Experimental Setup.}
We focus on instruction-guided image generation and propose enhancing performance through a model library, which retrieves components based on user instructions using model checkpoints and adapters. Details on its construction are in Appendix~\ref{app:ID}.
We compare our approach with two fine-tuning methods: InstructP2P, which adapts Stable Diffusion with small-scale instructional data, and Bagel, which combines large diffusion models with LLM supervision. These baselines allow evaluation against both lightweight and high-resource fine-tuning methods.

\begin{figure*}[!h]
\centering
\includegraphics[width=0.93\linewidth]{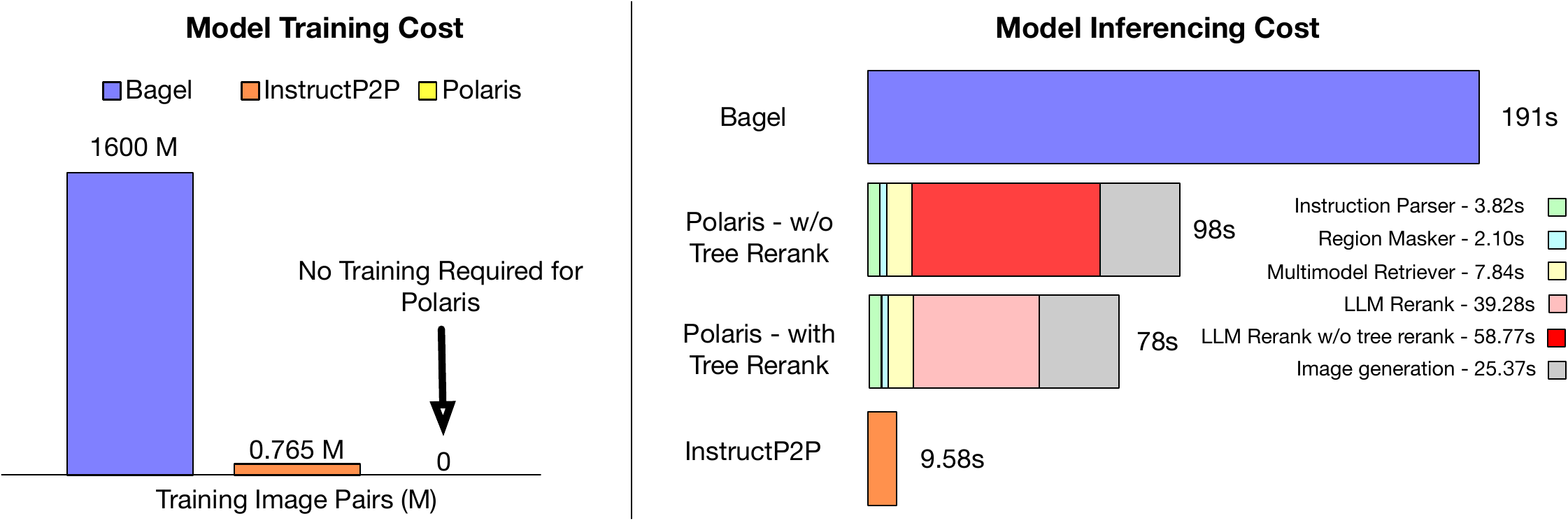}
\caption{
Comparison of training and inference efficiency across methods. Our approach, Polaris, requires no additional training and achieves inference time between the two baselines. Furthermore, by incorporating our proposed LLM Tree Rerank strategy, Polaris attains a 1.50× speedup in reranking, highlighting its practical efficiency advantage.}
\label{efficiency_test}
\vspace{-12pt}
\end{figure*}

\begin{figure*}[t]
\centering
\vspace{-3pt}
\includegraphics[width=0.98\linewidth]{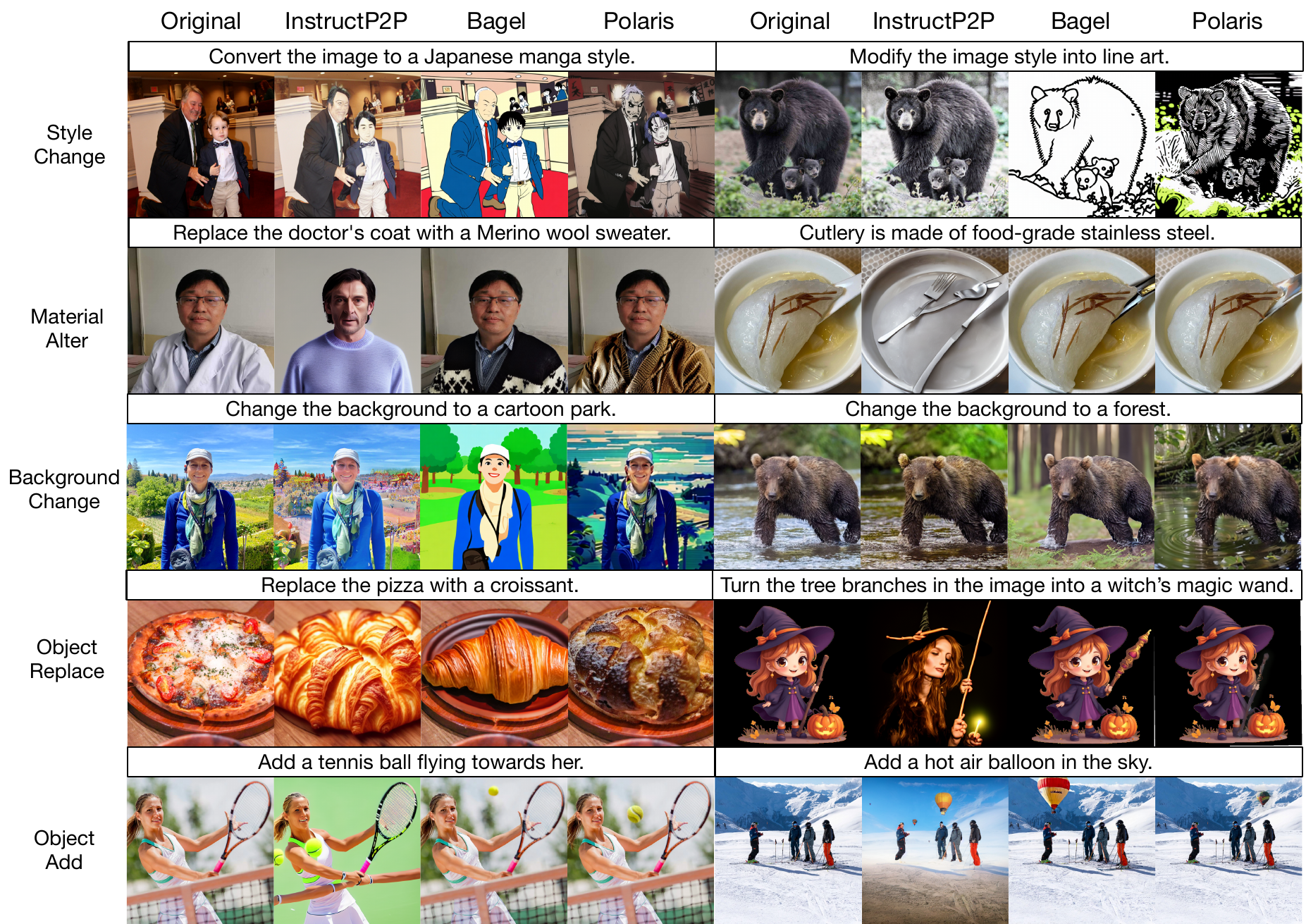}
\caption{Qualitative results on different tasks of \textsc{GEdit-Bench}~\citep{gedit}. Polaris achieves notably higher instruction-following success rates and more faithful outputs than InstructP2P, effectively handling style transformations that remain challenging even after fine-tuning. Moreover, the knowledge coverage of Polaris is comparable to Bagel, a trillion-token scale pretrained model.}
\label{common_result}
\vspace{-12pt}
\end{figure*}



\paragraph{Evaluation Protocol.}
To assess instruction-following ability, we adopt a subset of tasks from \textsc{GEdit-Bench}~\citep{gedit}—including background change, style change, material alteration, subject replacement, and subject addition.
We focus on this subset because it contains style-related transformations, which form the natural intersection between our style-aware instruction setting and classical image editing.

To further evaluate model performance under diverse, real-world user demands, we construct a supplementary benchmark, User-Bench, derived from community usage. In practice, users frequently upload the outputs of models to online communities after applying them to their own prompts, thereby providing both the input (the user prompt) and the corresponding output (the generated image). Leveraging this process, we curate a natural test set that reflects authentic user demands, and conduct a series of evaluations based on this dataset. The dataset construction process is detailed in Appendix~\ref{app:DG}. We referred to \textsc{GEdit-Bench} and employed VLM as the evaluation tool. The prompts used for the GPT-4o~\citep{gpt4o} evaluation are provided in Appendix~ \ref{app:EP}.

\subsection{Evaluating Training Blind Spots in Image Generation}

We evaluate our method on blind spots—cases where existing finetuned models systematically fail to generate reliable outputs, despite massive training datasets. Such blind spots often arise from factors such as data imbalance or catastrophic forgetting. To address this challenge, our method leverages retrieval from a model library, avoiding additional finetuning while substantially improving performance. We categorize blind spots into four types: object blind spots, character blind spots, style blind spots, and functional blind spots. As shown in Figure~\ref{blind_spot}, our approach consistently outperforms the unified multi-modal baseline across all categories, producing more faithful and diverse results.


\subsection{Evaluation: Quantitative and Qualitative}
\paragraph{Quantitative Result.} 
On our User-Bench, we evaluate two tasks: Local Style Change and Style Extraction. Local Style Change, which involves editing within about 30 predefined styles, shows that our method outperforms InstructP2P and is comparable to Bagel, though its advantage is limited by the small and fixed style set. In contrast, Style Extraction treats each user request as a unique style, leading to a dramatic increase in style diversity and better reflecting real-world usage. As shown in Table~\ref{tab:ourbench_sub}, our method demonstrates clear superiority, particularly in rare or hard-to-describe styles, since our embedding-based style retrieval captures nuanced user-specific requirements that traditional text-prompt-driven models often miss. 

On GEdit-Bench (Table~\ref{tab:eval}), our framework delivers strong performance on common editing cases, achieving results on par with instruction-finetuned models such as InstructP2P. Notably, in scenarios involving substantial style transformations, our approach exhibits clear advantages, underscoring its effectiveness in handling complex edits. Although large unified multi-modal models achieve higher overall scores, our framework—built upon conventional Stable Diffusion—comes remarkably close, highlighting the competitiveness of our lightweight and modular design.

\paragraph{Qualitative Result.} 
In Figure~\ref{common_result}, we present qualitative comparisons on GEdit-Bench. On standard editing tests, our approach exhibits much stronger instruction following than InstructP2P, producing edits that are both faithful and visually coherent. Remarkably, in terms of visual quality, our results are comparable to those of Bagel, a trillion-token scale pretrained model. Additional qualitative results on User-Bench are provided in Appendix~\ref{app:additional}.

\subsection{Efficiency}

We evaluate efficiency from two perspectives: training cost and inference latency, as shown in Figure~\ref{efficiency_test}. Our method requires no additional training or task-specific finetuning, introducing \textbf{zero extra training overhead} on top of the existing model library.  In contrast, Bagel and InstructP2P require $\sim$1.6M and $\sim$765K image pairs, respectively, making our approach far more practical and deployment-ready without costly retraining.  By building upon the shared model library, the training cost is effectively distributed across the entire community, rather than concentrated on a single provider.

For inference, our method runs at $\sim$78s per image, faster than Bagel ($\sim$191s) though slower than InstructP2P ($\sim$9.5s). Moreover, by incorporating our LLM-based tree reranking strategy, we further achieve a \textbf{1.50x speedup} in the reranking stage. Considering the elimination of training cost, our approach achieves a highly favorable balance between efficiency and applicability.

\section{Conclusion}
We present Polaris, a retrieval-based framework that scales instruction-guided image generation using community models. With multimodal retrieval and efficient adapter selection, it better captures style intent while remaining efficient. This work shows the promise of retrieval-driven approaches for scalable, personalized generation.

\section*{Impact Statement}
This paper presents work whose goal is to advance the field of Machine Learning. There are many potential societal consequences in our work, none of which we feel must be specifically highlighted here.
\bibliography{example_paper}

\begin{thebibliography}{47}
\providecommand{\natexlab}[1]{#1}
\providecommand{\url}[1]{\texttt{#1}}
\expandafter\ifx\csname urlstyle\endcsname\relax
  \providecommand{\doi}[1]{doi: #1}\else
  \providecommand{\doi}{doi: \begingroup \urlstyle{rm}\Url}\fi

\bibitem[Bai et~al.(2025)Bai, Chen, Liu, Wang, Ge, Song, Dang, Wang, Wang,
  Tang, Zhong, Zhu, Yang, Li, Wan, Wang, Ding, Fu, Xu, Ye, Zhang, Xie, Cheng,
  Zhang, Yang, Xu, and Lin]{qwen25vl}
Bai, S., Chen, K., Liu, X., Wang, J., Ge, W., Song, S., Dang, K., Wang, P.,
  Wang, S., Tang, J., Zhong, H., Zhu, Y., Yang, M., Li, Z., Wan, J., Wang, P.,
  Ding, W., Fu, Z., Xu, Y., Ye, J., Zhang, X., Xie, T., Cheng, Z., Zhang, H.,
  Yang, Z., Xu, H., and Lin, J.
\newblock Qwen2.5-vl technical report, 2025.

\bibitem[Brooks et~al.(2023{\natexlab{a}})Brooks, Holynski, and Efros]{edit}
Brooks, T., Holynski, A., and Efros, A.~A.
\newblock Instructpix2pix: Learning to follow image editing instructions.
\newblock In \emph{Proceedings of the IEEE/CVF conference on computer vision
  and pattern recognition}, pp.\  18392--18402, 2023{\natexlab{a}}.

\bibitem[Brooks et~al.(2023{\natexlab{b}})Brooks, Holynski, and
  Efros]{instructp2p}
Brooks, T., Holynski, A., and Efros, A.~A.
\newblock Instructpix2pix: Learning to follow image editing instructions.
\newblock In \emph{Proceedings of the IEEE/CVF conference on computer vision
  and pattern recognition}, pp.\  18392--18402, 2023{\natexlab{b}}.

\bibitem[Esser et~al.(2024)Esser, Kulal, Blattmann, Entezari, M{\"u}ller,
  Saini, Levi, Lorenz, Sauer, Boesel, et~al.]{sd3}
Esser, P., Kulal, S., Blattmann, A., Entezari, R., M{\"u}ller, J., Saini, H.,
  Levi, Y., Lorenz, D., Sauer, A., Boesel, F., et~al.
\newblock Scaling rectified flow transformers for high-resolution image
  synthesis.
\newblock In \emph{Forty-first international conference on machine learning},
  2024.

\bibitem[Falk et~al.(2025)Falk, Meynent, Pfammatter, Sch{\"u}rholt, and
  Borth]{zoo_vit}
Falk, D., Meynent, L., Pfammatter, F., Sch{\"u}rholt, K., and Borth, D.
\newblock A model zoo of vision transformers.
\newblock \emph{arXiv preprint arXiv:2504.10231}, 2025.

\bibitem[Gao et~al.(2023)Gao, Liu, Zeng, Xu, Li, Luo, Liu, Zhen, and
  Zhang]{super}
Gao, S., Liu, X., Zeng, B., Xu, S., Li, Y., Luo, X., Liu, J., Zhen, X., and
  Zhang, B.
\newblock Implicit diffusion models for continuous super-resolution.
\newblock In \emph{Proceedings of the IEEE/CVF conference on computer vision
  and pattern recognition}, pp.\  10021--10030, 2023.

\bibitem[Goodfellow et~al.(2015)Goodfellow, Mirza, Xiao, Courville, and
  Bengio]{catastrophic_survey}
Goodfellow, I.~J., Mirza, M., Xiao, D., Courville, A., and Bengio, Y.
\newblock An empirical investigation of catastrophic forgetting in
  gradient-based neural networks.
\newblock \emph{arXiv preprint arXiv:1312.6211}, 2015.

\bibitem[Ha et~al.(2016)Ha, Dai, and Le]{hyper}
Ha, D., Dai, A., and Le, Q.~V.
\newblock Hypernetworks.
\newblock \emph{arXiv preprint arXiv:1609.09106}, 2016.

\bibitem[Han et~al.(2024)Han, Gao, Liu, Zhang, and
  Zhang]{para_efficient_survey}
Han, Z., Gao, C., Liu, J., Zhang, J., and Zhang, S.~Q.
\newblock Parameter-efficient fine-tuning for large models: A comprehensive
  survey, 2024.

\bibitem[He et~al.(2022)He, Zhou, Ma, Berg-Kirkpatrick, and
  Neubig]{view_para_efficient}
He, J., Zhou, C., Ma, X., Berg-Kirkpatrick, T., and Neubig, G.
\newblock Towards a unified view of parameter-efficient transfer learning.
\newblock \emph{arXiv preprint arXiv:2110.04366}, 2022.

\bibitem[Hessel et~al.(2021)Hessel, Holtzman, Forbes, Bras, and Choi]{clip}
Hessel, J., Holtzman, A., Forbes, M., Bras, R.~L., and Choi, Y.
\newblock Clipscore: A reference-free evaluation metric for image captioning.
\newblock \emph{arXiv preprint arXiv:2104.08718}, 2021.

\bibitem[Ho et~al.(2020)Ho, Jain, and Abbeel]{ddpm}
Ho, J., Jain, A., and Abbeel, P.
\newblock Denoising diffusion probabilistic models.
\newblock \emph{arXiv preprint arXiv:2006.11239}, 2020.

\bibitem[Hu et~al.(2022)Hu, Shen, Wallis, Allen-Zhu, Li, Wang, Wang, Chen,
  et~al.]{lora}
Hu, E.~J., Shen, Y., Wallis, P., Allen-Zhu, Z., Li, Y., Wang, S., Wang, L.,
  Chen, W., et~al.
\newblock Lora: Low-rank adaptation of large language models.
\newblock \emph{ICLR}, 1\penalty0 (2):\penalty0 3, 2022.

\bibitem[Huang et~al.(2023)Huang, Liu, Lin, Pang, Du, and Lin]{lorahub}
Huang, C., Liu, Q., Lin, B.~Y., Pang, T., Du, C., and Lin, M.
\newblock Lorahub: Efficient cross-task generalization via dynamic lora
  composition.
\newblock \emph{arXiv preprint arXiv:2307.13269}, 2023.

\bibitem[Jin et~al.(2023)Jin, Xu, Chen, Liao, Tan, Huang, Chen, Lei, Liu, Song,
  et~al.]{lavit}
Jin, Y., Xu, K., Chen, L., Liao, C., Tan, J., Huang, Q., Chen, B., Lei, C.,
  Liu, A., Song, C., et~al.
\newblock Unified language-vision pretraining in llm with dynamic discrete
  visual tokenization.
\newblock \emph{arXiv preprint arXiv:2309.04669}, 2023.

\bibitem[Johnson \& Khoshgoftaar(2019)Johnson and
  Khoshgoftaar]{class_imbalance_survey}
Johnson, J.~M. and Khoshgoftaar, T.~M.
\newblock Survey on deep learning with class imbalance.
\newblock \emph{Journal of Big Data}, 2019.

\bibitem[Kirillov et~al.(2023)Kirillov, Mintun, Ravi, Mao, Rolland, Gustafson,
  Xiao, Whitehead, Berg, Lo, Dollar, and Girshick]{sam}
Kirillov, A., Mintun, E., Ravi, N., Mao, H., Rolland, C., Gustafson, L., Xiao,
  T., Whitehead, S., Berg, A.~C., Lo, W.-Y., Dollar, P., and Girshick, R.
\newblock Segment anything.
\newblock In \emph{Proceedings of the IEEE/CVF International Conference on
  Computer Vision (ICCV)}, pp.\  4015--4026, October 2023.

\bibitem[Li et~al.(2024)Li, Zeng, Feng, Gao, Liu, Liu, Li, Tang, Hu, Liu,
  et~al.]{Zone}
Li, S., Zeng, B., Feng, Y., Gao, S., Liu, X., Liu, J., Li, L., Tang, X., Hu,
  Y., Liu, J., et~al.
\newblock Zone: Zero-shot instruction-guided local editing.
\newblock In \emph{Proceedings of the IEEE/CVF Conference on Computer Vision
  and Pattern Recognition}, pp.\  6254--6263, 2024.

\bibitem[Liang et~al.(2023)Liang, Wu, Dai, Li, Zhao, Zhang, Zhang, Vajda, and
  Marculescu]{ovseg}
Liang, F., Wu, B., Dai, X., Li, K., Zhao, Y., Zhang, H., Zhang, P., Vajda, P.,
  and Marculescu, D.
\newblock Open-vocabulary semantic segmentation with mask-adapted clip.
\newblock In \emph{Proceedings of the IEEE/CVF conference on computer vision
  and pattern recognition}, pp.\  7061--7070, 2023.

\bibitem[Liu et~al.(2025)Liu, Han, Xing, Yin, Wang, Cheng, Liao, Wang, Fu, Han,
  et~al.]{gedit}
Liu, S., Han, Y., Xing, P., Yin, F., Wang, R., Cheng, W., Liao, J., Wang, Y.,
  Fu, H., Han, C., et~al.
\newblock Step1x-edit: A practical framework for general image editing.
\newblock \emph{arXiv preprint arXiv:2504.17761}, 2025.

\bibitem[Loke et~al.(2025)Loke, Quadri, Vivanco, Casagrande, and
  Fenollosa]{overcome_catastrophic}
Loke, B. S.~Y., Quadri, F., Vivanco, G., Casagrande, M., and Fenollosa, S.
\newblock Overcoming catastrophic forgetting in neural networks.
\newblock \emph{arXiv preprint arXiv:2507.10485}, 2025.

\bibitem[Lopez-Paz \& Ranzato(2022)Lopez-Paz and Ranzato]{continual_learning}
Lopez-Paz, D. and Ranzato, M.
\newblock Gradient episodic memory for continual learning.
\newblock \emph{arXiv preprint arXiv:1706.08840}, 2022.

\bibitem[Luo et~al.(2024)Luo, Wong, Trabucco, Huang, Gonzalez, Salakhutdinov,
  Stoica, et~al.]{stylus}
Luo, M., Wong, J., Trabucco, B., Huang, Y., Gonzalez, J.~E., Salakhutdinov, R.,
  Stoica, I., et~al.
\newblock Stylus: Automatic adapter selection for diffusion models.
\newblock \emph{Advances in Neural Information Processing Systems},
  37:\penalty0 32888--32915, 2024.

\bibitem[Michelessa et~al.(2025)Michelessa, Ng, Hurter, and Lim]{Varif_ai}
Michelessa, M., Ng, J., Hurter, C., and Lim, B.~Y.
\newblock Varif.ai to vary and verify user-driven diversity in scalable image
  generation.
\newblock In \emph{Proceedings of the 2025 ACM Designing Interactive Systems
  Conference}, DIS ’25, pp.\  1867–1885. ACM, July 2025.
\newblock \doi{10.1145/3715336.3735847}.
\newblock URL \url{http://dx.doi.org/10.1145/3715336.3735847}.

\bibitem[OpenAI et~al.(2024)]{gpt4o}
OpenAI et~al.
\newblock Gpt-4o system card.
\newblock \emph{arXiv preprint arXiv:2410.21276}, 2024.

\bibitem[Podell et~al.(2023)Podell, English, Lacey, Blattmann, Dockhorn,
  M{\"u}ller, Penna, and Rombach]{sdxl}
Podell, D., English, Z., Lacey, K., Blattmann, A., Dockhorn, T., M{\"u}ller,
  J., Penna, J., and Rombach, R.
\newblock Sdxl: Improving latent diffusion models for high-resolution image
  synthesis.
\newblock \emph{arXiv preprint arXiv:2307.01952}, 2023.

\bibitem[Qi et~al.(2024)Qi, Fang, Wu, Xie, Liu, Chen, He, and Zhang]{transfer}
Qi, T., Fang, S., Wu, Y., Xie, H., Liu, J., Chen, L., He, Q., and Zhang, Y.
\newblock Deadiff: An efficient stylization diffusion model with disentangled
  representations.
\newblock In \emph{Proceedings of the IEEE/CVF conference on computer vision
  and pattern recognition}, pp.\  8693--8702, 2024.

\bibitem[Ramesh \& Chaudhari(2021)Ramesh and Chaudhari]{zoo_2021}
Ramesh, R. and Chaudhari, P.
\newblock Model zoo: A growing" brain" that learns continually.
\newblock \emph{arXiv preprint arXiv:2106.03027}, 2021.

\bibitem[Rombach et~al.(2022)Rombach, Blattmann, Lorenz, Esser, and Ommer]{ldm}
Rombach, R., Blattmann, A., Lorenz, D., Esser, P., and Ommer, B.
\newblock High-resolution image synthesis with latent diffusion models.
\newblock In \emph{Proceedings of the IEEE/CVF conference on computer vision
  and pattern recognition}, pp.\  10684--10695, 2022.

\bibitem[Ruiz et~al.(2023{\natexlab{a}})Ruiz, Li, Jampani, Pritch, Rubinstein,
  and Aberman]{dreambooth}
Ruiz, N., Li, Y., Jampani, V., Pritch, Y., Rubinstein, M., and Aberman, K.
\newblock Dreambooth: Fine tuning text-to-image diffusion models for
  subject-driven generation.
\newblock In \emph{Proceedings of the IEEE/CVF conference on computer vision
  and pattern recognition}, pp.\  22500--22510, 2023{\natexlab{a}}.

\bibitem[Ruiz et~al.(2023{\natexlab{b}})Ruiz, Li, Jampani, Pritch, Rubinstein,
  and Aberman]{finetune}
Ruiz, N., Li, Y., Jampani, V., Pritch, Y., Rubinstein, M., and Aberman, K.
\newblock Dreambooth: Fine tuning text-to-image diffusion models for
  subject-driven generation.
\newblock In \emph{Proceedings of the IEEE/CVF conference on computer vision
  and pattern recognition}, pp.\  22500--22510, 2023{\natexlab{b}}.

\bibitem[Saharia et~al.(2022)Saharia, Chan, Chang, Lee, Ho, Salimans, Fleet,
  and Norouzi]{complete}
Saharia, C., Chan, W., Chang, H., Lee, C., Ho, J., Salimans, T., Fleet, D., and
  Norouzi, M.
\newblock Palette: Image-to-image diffusion models.
\newblock In \emph{ACM SIGGRAPH 2022 conference proceedings}, pp.\  1--10,
  2022.

\bibitem[Shi et~al.(2024)Shi, Han, Zhou, Liang, Lin, Zettlemoyer, and
  Yu]{lmfusion}
Shi, W., Han, X., Zhou, C., Liang, W., Lin, X.~V., Zettlemoyer, L., and Yu, L.
\newblock Lmfusion: Adapting pretrained language models for multimodal
  generation.
\newblock \emph{arXiv preprint arXiv:2412.15188}, 2024.

\bibitem[Singh(2023)]{t2i_t2v_survey}
Singh, A.
\newblock A survey of ai text-to-image and ai text-to-video generators.
\newblock In \emph{2023 4th International Conference on Artificial
  Intelligence, Robotics and Control (AIRC)}, pp.\  32–36. IEEE, May 2023.
\newblock \doi{10.1109/airc57904.2023.10303174}.

\bibitem[Song et~al.(2020)Song, Sohl-Dickstein, Kingma, Kumar, Ermon, and
  Poole]{diffusion}
Song, Y., Sohl-Dickstein, J., Kingma, D.~P., Kumar, A., Ermon, S., and Poole,
  B.
\newblock Score-based generative modeling through stochastic differential
  equations.
\newblock \emph{arXiv preprint arXiv:2011.13456}, 2020.

\bibitem[Song et~al.(2021)Song, Sohl-Dickstein, Kingma, Kumar, Ermon, and
  Poole]{score_based_generative}
Song, Y., Sohl-Dickstein, J., Kingma, D.~P., Kumar, A., Ermon, S., and Poole,
  B.
\newblock Score-based generative modeling through stochastic differential
  equations, 2021.

\bibitem[Sun et~al.(2023)Sun, Yu, Cui, Zhang, Zhang, Wang, Gao, Liu, Huang, and
  Wang]{emu}
Sun, Q., Yu, Q., Cui, Y., Zhang, F., Zhang, X., Wang, Y., Gao, H., Liu, J.,
  Huang, T., and Wang, X.
\newblock Emu: Generative pretraining in multimodality.
\newblock \emph{arXiv preprint arXiv:2307.05222}, 2023.

\bibitem[Team(2024)]{chameleon}
Team, C.
\newblock Chameleon: Mixed-modal early-fusion foundation models.
\newblock \emph{arXiv preprint arXiv:2405.09818}, 2024.

\bibitem[Tian et~al.(2024)Tian, Aggarwal, Colaco, Kira, and
  Gonzalez-Franco]{segment}
Tian, J., Aggarwal, L., Colaco, A., Kira, Z., and Gonzalez-Franco, M.
\newblock Diffuse attend and segment: Unsupervised zero-shot segmentation using
  stable diffusion.
\newblock In \emph{Proceedings of the IEEE/CVF Conference on Computer Vision
  and Pattern Recognition}, pp.\  3554--3563, 2024.

\bibitem[Wei et~al.(2025)Wei, Zheng, Zhang, Liu, Ji, Zhang, and
  Zuo]{personalized_img_gen_sursey}
Wei, Y., Zheng, Y., Zhang, Y., Liu, M., Ji, Z., Zhang, L., and Zuo, W.
\newblock Personalized image generation with deep generative models: A decade
  survey.
\newblock \emph{arXiv preprint arXiv:2502.13081}, 2025.

\bibitem[Yang et~al.(2023)Yang, Zhang, Song, Hong, Xu, Zhao, Zhang, Cui, and
  Yang]{diffusion_survey}
Yang, L., Zhang, Z., Song, Y., Hong, S., Xu, R., Zhao, Y., Zhang, W., Cui, B.,
  and Yang, M.-H.
\newblock Diffusion models: A comprehensive survey of methods and applications.
\newblock \emph{ACM computing surveys}, 56\penalty0 (4):\penalty0 1--39, 2023.

\bibitem[Yin et~al.(2024)Yin, Fu, Zhao, Li, Sun, Xu, and Chen]{mllm_survey}
Yin, S., Fu, C., Zhao, S., Li, K., Sun, X., Xu, T., and Chen, E.
\newblock A survey on multimodal large language models.
\newblock \emph{National Science Review}, 11\penalty0 (12), November 2024.
\newblock ISSN 2053-714X.
\newblock \doi{10.1093/nsr/nwae403}.

\bibitem[Zhang et~al.(2024)Zhang, Yu, Dong, Li, Su, Chu, and Yu]{mmllms_survey}
Zhang, D., Yu, Y., Dong, J., Li, C., Su, D., Chu, C., and Yu, D.
\newblock Mm-llms: Recent advances in multimodal large language models.
\newblock \emph{arXiv preprint arXiv:2401.13601}, 2024.

\bibitem[Zhang et~al.(2023)Zhang, Wang, Huang, Tasnim, and
  Shi]{diff_base_img_gen_survey}
Zhang, T., Wang, Z., Huang, J., Tasnim, M.~M., and Shi, W.
\newblock A survey of diffusion based image generation models: Issues and their
  solutions.
\newblock \emph{arXiv preprint arXiv:2308.13142}, 2023.

\bibitem[Zhang et~al.(2025)Zhang, Guo, Zhao, Fu, Duan, Hu, Chng, Wang, Chen,
  Xu, Luo, and Zhang]{unified_survey}
Zhang, X., Guo, J., Zhao, S., Fu, M., Duan, L., Hu, J., Chng, Y.~X., Wang,
  G.-H., Chen, Q.-G., Xu, Z., Luo, W., and Zhang, K.
\newblock Unified multimodal understanding and generation models: Advances,
  challenges, and opportunities, 2025.

\bibitem[Zhao et~al.(2024)Zhao, Gan, Wang, Zhou, Yang, Kuang, and
  Wu]{loraretriever}
Zhao, Z., Gan, L., Wang, G., Zhou, W., Yang, H., Kuang, K., and Wu, F.
\newblock Loraretriever: Input-aware lora retrieval and composition for mixed
  tasks in the wild.
\newblock \emph{arXiv preprint arXiv:2402.09997}, 2024.

\bibitem[Zhou et~al.(2024)Zhou, Yu, Babu, Tirumala, Yasunaga, Shamis, Kahn, Ma,
  Zettlemoyer, and Levy]{transfusion}
Zhou, C., Yu, L., Babu, A., Tirumala, K., Yasunaga, M., Shamis, L., Kahn, J.,
  Ma, X., Zettlemoyer, L., and Levy, O.
\newblock Transfusion: Predict the next token and diffuse images with one
  multi-modal model.
\newblock \emph{arXiv preprint arXiv:2408.11039}, 2024.

\end{thebibliography}
\bibliographystyle{icml2026}

\onecolumn

\appendix

\section{Additional Experiment Results}\label{app:additional}
\begin{figure*}[!ht]
\centering
\includegraphics[width=\linewidth]{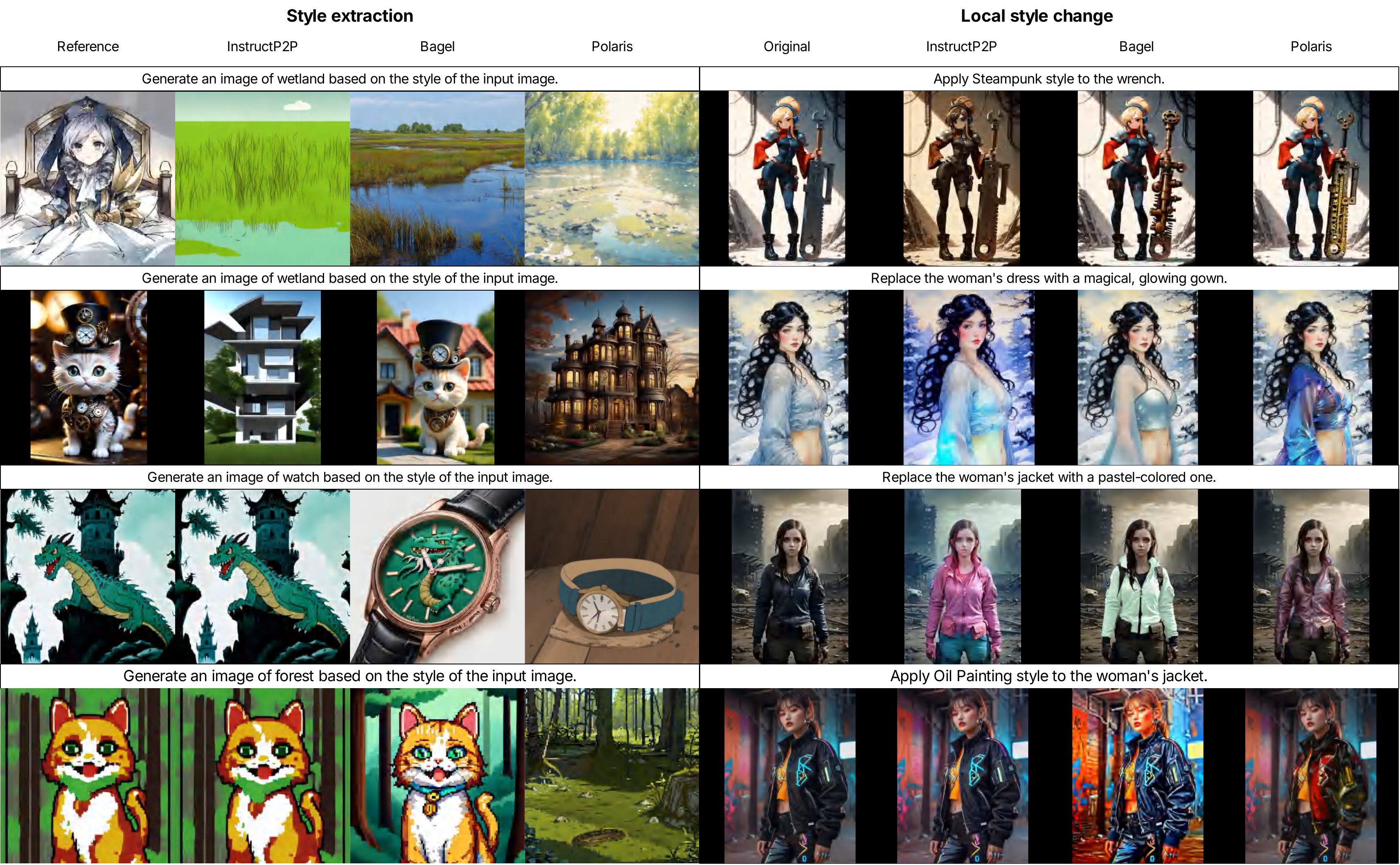}
\caption{Experimental results on User-Bench. In the style extraction task, our method outperforms both baselines in terms of instruction understanding and generation quality. In the local style change task, our results surpass InstructP2P and are comparable to those of Bagel.}
\label{additional}
\vspace{-10pt}
\end{figure*}

Figure ~\ref{additional} provides additional qualitative results on the User-Bench benchmark.
For the style extraction task, our method produces outputs with clearer adherence to the input instructions and more consistent stylistic fidelity compared to both baselines.
For the local style change task, our approach delivers more targeted edits while preserving the surrounding content, outperforming InstructP2P and achieving results comparable to Bagel.
These examples further illustrate the model’s ability to handle user-driven stylistic transformations under different editing scenarios.

\section{Implementation Details}\label{app:ID}
We use Stable Diffusion 1.5 (SD 1.5) as the baseline generative model for all experiments, as it is the most widely adopted open-source backbone with extensive community-contributed adapters. We collect models from the Civitai platform, excluding those containing NSFW content. To build our retrieval system, we construct a model library comprising approximately 6,700 community-shared checkpoints, each representing a user-uploaded variant of SD 1.5, together with about 75,000 LoRA adapters (split following~\citep{stylus}). Among these adapters, roughly 9,000 are dedicated to style generation and 65,000 to object generation, each paired with embeddings that encode their style or task specifications. For reranking retrieved candidates, we employ Qwen2.5-14B as a large language model (LLM) reranker, which improves alignment between user instructions and the selected checkpoints or LoRAs. Both the base diffusion model and all variants in the database are implemented within the SD 1.5 framework, ensuring compatibility and controllability across the system.

\section{Method Details}
\subsection{Instruction Parser}\label{app:IP}
To interpret user inputs, we employ a vision–language model (VLM) guided by a tailored prompt. The parser analyzes the instruction along several dimensions: \textit{instruction type} (categorizing the input task), \textit{style transformation} (whether a style change is required), \textit{subject} (the target entity of modification), \textit{generation prompt} (translating instructions into standard prompts since SD 1.5 does not support direct instruction inputs), \textit{contour compatibility} (whether large-scale shape changes are involved), and \textit{foreground elements} (which foreground objects are to be edited). The full prompt design is provided in Table~\ref{tab:analyst-prompt}.

\begin{table*}[h!]
\centering
\scriptsize
\renewcommand{\arraystretch}{1.15}
\setlength{\tabcolsep}{5pt}
\begin{tabular}{@{}p{0.95\textwidth}@{}}
\toprule
\textbf{Given a user's image editing request in natural language, extract the following elements:} \\
\midrule
\textbf{1. "instruction\_type": Identify the editing intent.} \\[3pt]
- 0 = Category change \newline
\quad • Condition: The target object's core category \textbf{changes to a different class}. \newline
\quad Decide \textbf{only} by category, \textbf{not} by verbs like “replace/change”. \newline
\quad If A and B share the same high-level category, do \textbf{not} use 0. \newline
\quad • Examples: "change the cat to a dog", "replace a chair with a table" \newline
\quad • Note: If the target category stays the same but only style/appearance changes \newline
\quad \ \ \ \ (e.g., "replace the cat by a watercolor cat"), use 2 instead. \\[3pt]
- 1 = Add new object \newline
\quad • Condition: A new object/element is introduced into the scene. \newline
\quad • Examples: "add a hat to the person", "put a bird in the sky", "add a tree next to the house" \\[3pt]
- 2 = Local style/appearance transformation \newline
\quad • Condition: The object category remains the same, but its style/appearance changes. \newline
\quad • Examples: "make the cloth black and white", "turn the cat into a watercolor cat", "change the shoe into a red shoe" \newline
\quad • Special case: If the modification explicitly targets the "background" (e.g., "adjust the background to a forest", "make the background blurry"), treat it as type 2, since "background" is a specific region rather than the whole image. \\[3pt]
- 3 = Global style/appearance transformation \newline
\quad • Condition: A style/effect is applied to the entire image. \newline
\quad • Examples: "make the image black and white", "apply a watercolor filter to the whole picture" \\[3pt]
- 4 = Style transfer from reference image \newline
\quad • Condition: The user provides a reference image and requests generating new content \newline
\quad \ \ \ \ (objects, scenes, characters, etc.) \textbf{in the style of that reference image}. \newline
\quad • Key difference: Unlike 0--3, this does \textbf{not} edit the original image; the reference serves only as a style source. \newline
\quad • Examples: "generate a cat in the style of the reference image", "create a cityscape with the style from the given artwork". \\
\midrule
\textbf{2. "style\_transformation": Binary flag (0 or 1).} \\
- Relevant for instruction\_type = 0 or 1. \newline
- 1 = The request explicitly specifies a style or artistic effect. \newline
\quad • Example: "turn the cat into a watercolor dog" → 1 \newline
\quad • Example: "add a watercolor cat" → 1 \newline
- 0 = No style/effect is mentioned. \newline
\quad • Example: "change the cat to a dog" → 0 \newline
\quad • Example: "add a cat" → 0 \newline
- For instruction\_type = 2 or 3, this value is always 1 by definition. \\
\midrule
\textbf{3. "subject"} \\
Must strictly follow this structure → "\textless TARGET\_TYPE\textgreater + \textless LOCAL\_RANGE\textgreater". \newline
- \textless TARGET\_TYPE\textgreater: The core object name, no possessive forms. \newline
\quad • Correct: "cloth", "shoe" \newline
\quad • Incorrect: "girl's cloth", "man's shoe" \newline
- \textless LOCAL\_RANGE\textgreater: Spatial area where the target is located. \newline
\quad • Examples: "on the upper body", "in the foreground", "near the tree" \\
\midrule
\textbf{4. "generation\_prompt"} \\
Describe only the \textbf{final appearance} of the object/region. \newline
- Do NOT mention the editing action (replace/change/remove). \newline
- Do NOT add attributes beyond the request. \newline
- Keep it concise, faithful, and specific. \\
\midrule
\textbf{5. "contour\_compatibility": Strict binary flag (0 or 1).} \\
- 0 = Contour differs significantly. \newline
\quad • Examples: "flag → hat", "book → vase" \newline
- 1 = Contour is compatible. \newline
\quad • Examples: "apple → orange", "car → bus" \\
\midrule
\textbf{6. "foreground\_elements"} \\
A JSON array listing the distinct foreground objects in the image request. \newline
- Each element should be a simple noun phrase (e.g., "dog", "tree", "man", "hat"). \newline
- Keep them concise, no extra adjectives unless explicitly requested. \newline
- MUST NOT be empty. If uncertain, infer plausible main foreground objects from the request context. \\
\midrule
\textbf{Respond strictly in the following JSON format:} \\
\texttt{\{ } \\
\hspace*{1em}\texttt{"instruction\_type": <0|1|2|3|4>,} \\
\hspace*{1em}\texttt{"style\_transformation": <0|1>,} \\
\hspace*{1em}\texttt{"subject": "<specific image region or object to edit>",} \\
\hspace*{1em}\texttt{"generation\_prompt": "<final desired description>",} \\
\hspace*{1em}\texttt{"contour\_compatibility": <0|1>,} \\
\hspace*{1em}\texttt{"foreground\_elements": ["<object1>", "<object2>", ...]} \\
\texttt{\}} \\
\midrule
\textbf{User's request:} \texttt{"\{prompt\}"} \\
\bottomrule
\end{tabular}
\caption{Analyst prompt specification. The table outlines the schema and annotation rules used to extract structured representations, including instruction type, style transformation, subject, generation prompt, contour compatibility, and foreground elements.}

\label{tab:analyst-prompt}
\end{table*}

\subsection{Adapter Editor}\label{app:AE}
During image generation, we employ three types of adapters to influence the final output: \textit{style checkpoints}, \textit{style LoRAs}, and \textit{object LoRAs}. Depending on the instruction, different modalities are used for retrieval. For instructions that do not involve style transformation, we aim to preserve the original style of the edited image. In this case, a style checkpoint and a style LoRA are retrieved using image embeddings, while the object LoRA is retrieved using text embeddings. For instructions that explicitly involve style transformation, the original style is not preserved; instead, the style checkpoint, style LoRA, and object LoRA are all retrieved using text embeddings.

\section{Dataset Generation}\label{app:DG}
The evaluation dataset is constructed from user-uploaded images on Civitai, covering generations produced by diverse models. To ensure fair testing, we specifically select outputs generated by models with architectures different from SD 1.5. This structural difference keeps the test data disjoint from both the training data of the baseline model and the retrieval database, ensuring that neither the base model nor the retrieved checkpoints and LoRAs have been exposed to the evaluation samples. 

We design two types of test cases: \textit{style extraction} and \textit{local style change}.
For the \emph{local style change} task, we employ \texttt{Qwen2.5-VL-7B} to automatically generate prompts conditioned on the downloaded images. 
The complete prompt design is summarized in Table~\ref{tab:lm-prompt}.
For the \emph{style extraction} task, we first construct a subject pool and then randomly sample one subject. 
The resulting textual instruction is formulated as: ``Generate an image of <subject> based on the style of the input image.'' 

\begin{table*}[h!]
\centering
\scriptsize
\renewcommand{\arraystretch}{1.1}
\setlength{\tabcolsep}{5pt}
\begin{tabular}{@{}p{0.95\textwidth}@{}}
\toprule
\textbf{You are given an input image.} \\
The target visual style (S) for this edit is: \texttt{"\{style\}"}. You must apply this style as instructed below. Please follow these instructions carefully: \\
\midrule
\textbf{1. First, Identify and enumerate ALL main subjects in the image.} \\
A 'subject' refers to any physically distinct person, animal, object, background/scene, or any major visible part of a person or animal, such as hair (hairstyle), face, upper body clothing (shirt, jacket, dress), or lower body clothing (pants, skirt, dress). Do not include minor details like shoes, socks, glasses, or small accessories as separate subjects.\\
- For any person or animal, list major visible parts as separate subjects if they are visually distinct (for example: (1) person's hair, (2) person's face, (3) person's shirt, (4) person's pants, etc.). \\
- For any object that a person or animal is interacting with (e.g., a guitar being played, a book being held), also include it as a separate subject. \\
- Do NOT combine multiple items or persons as one subject. List each main subject and each major visible part separately. \\
- Example list: \newline
  (1) Woman's hair \newline
  (2) Woman's face \newline
  (3) Woman's dress \newline
  (4) Guitar (being played by the woman) \newline
  (5) Microphone \newline
  (6) The stage background \\
\midrule
\textbf{2.Count the number of subjects you listed, and use this as 'num\_subjects'.} \\
 \\
\midrule
\textbf{3. If \texttt{num\_subjects} \textgreater 1:} \\
- Randomly select \textbf{only one} subject to edit (for example, choose subject (4): Guitar). \\
- Replace the selected subject with a different object, entity, or new scene. \\
- Apply the style \texttt{"\{style\}"} ONLY to the replaced/new subject. Do NOT apply this style to the entire image. All other subjects and the rest of the image should keep their original style and appearance. \\
- Clearly state which subject (by number and description) was chosen in your edit\_instruction. \\
\midrule
\textbf{4. If \texttt{num\_subjects} = 1:} \\
- Do NOT replace the subject. Only apply the style \"{style}\" to the subject (and its background if the subject is the background itself). \\
\midrule
\textbf{5. Under no circumstances should you replace or modify more than one subject at a time. Do NOT apply the style globally.} \\
\\
\midrule
\textbf{6. Your response must be a JSON object using the following format:} \\

\texttt{\{} \\
\hspace*{1em}\texttt{"subjects": [ <a list of identified subjects as strings> ],} \\
\hspace*{1em}\texttt{"num\_subjects": <int>,} \\
\hspace*{1em}\texttt{"edit\_instruction": "<one concise sentence describing the single subject replacement (if any) and the style change, clearly stating the chosen subject by number and description>",} \\
\hspace*{1em}\texttt{"result\_prompt": "<a detailed description of the final image after editing, focusing on what is visually present in the image. The description should not mention any editing actions or changes. Only describe what can be directly seen in the resulting image.>"} \\
\texttt{\}} \\
\midrule
\textbf{Do NOT include markdown, code fences, or commentary — return only the JSON object.} \\
\bottomrule
\end{tabular}
\caption{Prompt template provided to the VLM during dataset construction. 
It specifies how the model should enumerate image subjects, select a single editing target, apply the designated visual style locally, and output a structured JSON object representing the generated image editing instruction.}

\label{tab:lm-prompt}
\end{table*}

\section{Evaluation Prompt}\label{app:EP}
We use GPT-5o-mini to evaluate the final results on User-Bench. For the two tasks, \textit{Local Style Change} and \textit{Style Extraction}, we design different evaluation prompts to better capture task-specific objectives. The prompt used for assessing style transformation in image editing is summarized in Table~\ref{tab:verification-prompt}.

\begin{table*}[h!]
\centering
\scriptsize
\renewcommand{\arraystretch}{1.15}
\setlength{\tabcolsep}{5pt}
\begin{tabular}{@{}p{0.95\textwidth}@{}}
\toprule
\textbf{System / Judge Instruction} \\
You are an image evaluation model. The evaluation target is \textbf{“Image Editing with Style Transformation.”} \\[3pt]
\textbf{Important Instruction:} You \textbf{must} always return a result, even if it’s not perfect. Ensure that you provide the requested evaluation for the image modification, including the scores and the reasons. \\
\midrule
\textbf{Task Definition} \\
The user provides a source image (with the object/region to be edited) and a text description.  
The model must keep the overall structure of the source image while \textbf{modifying the specified object/region}, transforming it into a new form and/or applying a new style. \\
\midrule
\textbf{Inputs} \\
- Source image: \textless SRC\_IMAGE \textgreater \newline
- Candidate image: \textless CAND\_IMAGE \textgreater \newline
- User text prompt: \textless USER\_TEXT \textgreater \\
\midrule
\textbf{Evaluation Focus} \\
- \textbf{Style integration}: Does the new style of the modified part appear consistent and well integrated with the whole image? \newline
- \textbf{Structural consistency}: Are unmodified regions preserved without unnecessary changes or corruption? \newline
- \textbf{Image quality}: Is the generated image visually clean, stable, and free of major flaws? \newline
- \textbf{actual\_modification}: To what extent have real, meaningful modifications been made to the image? \\
\midrule
\textbf{Scoring Rubric (Scores 0–4)} \\[3pt]
\begin{tabular}{p{0.45\textwidth} p{0.45\textwidth}}
\textbf{Style integration} & \textbf{Structural consistency} \\
0: Style completely wrong \newline
1: Slightly aligned \newline
2: Some correct, poor integration \newline
3: Largely consistent \newline
4: Highly consistent & 
0: Severe redraw/corruption \newline
1: Most areas degraded \newline
2: Majority preserved, issues \newline
3: Largely preserved \newline
4: Fully preserved \\
\midrule
\textbf{Image quality} & \textbf{Actual modification} \\
0: Severe artifacts \newline
1: Major flaws \newline
2: Acceptable \newline
3: Good quality \newline
4: Polished &
0: No real modification \newline
1: Minor tweaks only \newline
2: Substantial change \newline
3: Significant changes \newline
4: Major mods, intact structure \\

\end{tabular} \\

\midrule
\textbf{Output JSON Schema} \\
\texttt{\{ } \\
\hspace*{1em}\texttt{"style\_integration": 0-4,} \\
\hspace*{1em}\texttt{"structural\_consistency": 0-4,} \\
\hspace*{1em}\texttt{"image\_quality": 0-4,} \\
\hspace*{1em}\texttt{"actual\_modification": 0-4,} \\
\hspace*{1em}\texttt{"reasons": \{ } \\
\hspace*{2em}\texttt{"style": "< <=40 words>",} \\
\hspace*{2em}\texttt{"structure": "< <=40 words>",} \\
\hspace*{2em}\texttt{"quality": "< <=40 words>",} \\
\hspace*{2em}\texttt{"modification": "< <=40 words>",} \\
\hspace*{1em}\texttt{\}} \\
\texttt{\}} \\
\bottomrule
\end{tabular}
\caption{Evaluation prompt specification. The prompt is fed to GPT-5o-mini to verify style-transformed image edits. It defines the evaluation focus, scoring rubric, and the structured JSON schema required for standardized reporting of results.}

\label{tab:verification-prompt}
\end{table*}

\section{LLM Usage}
Large language models (LLMs) were used only for minor language editing and polishing of the manuscript. They were not involved in the design of the research, development of methods, execution of experiments, analysis of results, or generation of scientific content. The authors take full responsibility for the final content of the paper.

\end{document}